\pdfoutput=1
\documentclass[11pt]{article}

\usepackage[final]{acl}

\usepackage{times}
\usepackage{latexsym}

\usepackage[T1]{fontenc}
\usepackage[utf8]{inputenc}

\usepackage{microtype}

\usepackage{inconsolata}

\usepackage{graphicx}
\usepackage{multirow}
\usepackage{array}
\usepackage[table,xcdraw]{xcolor}
\usepackage{xcolor}
\usepackage{amssymb}
\usepackage{colortbl}
\usepackage{booktabs}
\usepackage{amsmath}
\title{Fine-Tuned Thoughts: Leveraging Chain-of-Thought Reasoning for Industrial Asset Health Monitoring}

\author{%
Shuxin Lin$^{1}$ \quad Dhaval Patel$^{1}$ \quad Christodoulos Constantinides$^{2}$ \\
$^1$IBM Research \quad $^2$IBM \\
\texttt{\{shuxin.lin@, pateldha@us., christodoulos.constantinides@\}ibm.com}}

\begin{document}
\maketitle
\begin{abstract}
Small Language Models (SLMs) are becoming increasingly popular in specialized fields, such as industrial applications, due to their efficiency, lower computational requirements, and ability to be fine-tuned for domain-specific tasks, enabling accurate and cost-effective solutions. However, performing complex reasoning using SLMs in specialized fields such as Industry 4.0 remains challenging. In this paper, we propose a knowledge distillation framework for industrial asset health, which transfers reasoning capabilities via Chain-of-Thought (CoT) distillation from Large Language Models (LLMs) to smaller, more efficient models (SLMs). We discuss the advantages and the process of distilling LLMs using multi-choice question answering (MCQA) prompts to enhance reasoning and refine decision-making. We also perform in-context learning to verify the quality of the generated knowledge and benchmark the performance of fine-tuned SLMs with generated knowledge against widely used LLMs. The results show that the fine-tuned SLMs with CoT reasoning outperform the base models by a significant margin, narrowing the gap to their LLM counterparts. Our code is open-sourced at: \href{https://github.com/IBM/FailureSensorIQ}{https://github.com/IBM/FailureSensorIQ}.
\end{abstract}

\section{Introduction}

Large Language Models (LLMs) have demonstrated exceptional proficiency in both generic and specialized domains due to their extensive pretraining on vast amounts of text data from diverse sources, that enables strong contextual understanding and reasoning. 
Small Language Models (SLMs), on the other hand, while perform well in common NLP tasks \citep{wang2024comprehensivesurveysmalllanguage} such as text classification, sentiment analysis, their limited parameter capacity ($\leq$8B) constrains their ability to store extensive knowledge and perform complex reasoning, making them less effective for specialized domains without substantial modifications.

Recent studies on Knowledge Distillation from industries and academics have shown that Small Language Models (SLMs) hold great potential in specialized domains, such as math reasoning \cite{shridhar-etal-2023-distilling}. The reduced computational requirements of SLMs allow for faster inference and deployment on resource-constrained devices. Also, SLMs can be fine-tuned, hosted and operated locally on computing machines, minimizing the need for sensitive user data and domain information to be exposed or leaked.  The benefits of SLMs bring opportunities to the manufacturing industries, including maintenance and monitoring, process optimization, and quality control. 

\begin{table}[hb!]
    \centering
    \scalebox{0.85}{
    \begin{tabular}{|>{}l|c|c|c|c|c|}
        \hline
        \multirow{2}{*}{\textcolor{black}{\textbf{Failure Mode}}} & \multicolumn{5}{c|}{\textbf{Sensor/Parameter Reading}} \\ 
        \cline{2-6}
        & \rotatebox{70}{Power} & \rotatebox{70}{Speed} & \rotatebox{70}{Pressure} & \rotatebox{70}{Vibr.} & \rotatebox{70}{Temp.} \\ 
        \hline
        Bearing wear & & \cellcolor{yellow!50}\checkmark & \cellcolor{yellow!50}\checkmark & & \cellcolor{yellow!50}\checkmark \\ 
        \hline
        Gear Defect & & & \cellcolor{red!40}\checkmark & \cellcolor{red!40}\checkmark & \\ 
        \hline
        Unbalance & \cellcolor{blue!40}\checkmark & & & & \cellcolor{blue!40}\checkmark \\ 
        \hline
        Shaft Misalignment & \cellcolor{green!50}\checkmark & \cellcolor{green!50}\checkmark & & \cellcolor{green!50}\checkmark & \\ 
        \hline
        Overheating & & & \cellcolor{purple!40}\checkmark & & \cellcolor{purple!40}\checkmark \\ 
        \hline
    \end{tabular}
    }
     \caption{Expert Knowledge: Failure Faults $\leftrightarrow$ Sensors/Parameters:  \checkmark indicates that parameter or sensor change if failure occurs}   
    \label{tab:mapping-table}
\end{table}

In this paper, we focus on the adoption of small language models in Industrial Asset Health applications, which involve monitoring the health of assets using sensor data. Typically, Internet of Things (IoT) devices collect data from a variety of sensors, including those that measure temperature, power, and pressure. The sensor data is then analyzed to predict potential failures, such as ``overheating'', and to recommend proactive maintenance before breakdowns occur. To improve failure detection, \textbf{Failure Modes and Effects Analysis (FMEA)} used in reliability engineering is commonly applied to both sensor data and failure modes (see Table \ref{tab:mapping-table}). This method establishes connections between an asset's potential failures and the monitored sensors that can signal these failures when anomalies are detected. Such analysis is a core component of industrial asset health monitoring. 

\begin{table}[t!]
    \centering
    \renewcommand{\arraystretch}{1} % Adjust row spacing
    \small
    
    \begin{tabular}{p{2cm} p{4.8cm}}   
        \toprule
        \textbf{Question Category} & \textbf{Example Question} \\
        \midrule
        Asset to Sensors & What are the sensors that could be useful in monitoring the condition of an asset? \\\midrule
        Failure Mode to Class & Given a failure description, which failure mode class does it belong to? \\\midrule
        Failure Mode to Sensor & To prevent an occurrence of a failure, what are the sensors that can be used to detect it early? \\\midrule
        Sensor to Failure Mode & When anomalies are detected in a sensor reading, what failure modes can be the root cause? \\ 
        \bottomrule
    \end{tabular}
    \caption{Example FMEA Questions by Category in Asset Health Monitoring Application}
    \label{tab:FMEA_questions}
\end{table}

Can a Large Language Model (LLM) act as a potential knowledge generator (see Table \ref{tab:mapping-table}), offering insights into the relationships between failure modes and sensor parameters? An LLM-based workflow could enhance decision-making by providing contextual understanding and reasoning for FMEA-related questions, as outlined in Table \ref{tab:FMEA_questions}.

\subsection{SLM Challenges for Asset Health Monitoring}
\textbf{Scarcity of high-quality, labeled data}. The general guidelines of FMEA in industrial asset maintenance was published by ISO Standards\footnote{\href{https://www.iso.org/standard/71194.html}{https://www.iso.org/standard/71194.html}} that cover only 10 assets. For emerging technologies or newer machinery that hasn't yet gone through extensive operational lifecycles, there may be limited data or documents on failure modes. This scarcity makes it difficult to perform a thorough FMEA and predict potential failures accurately, as historical failure data simply doesn't recorded yet.

\textbf{Genericness of LLM response to a specific domain}. LLMs are trained on vast, general datasets that cover a wide range of topics. A domain-specific context (e.g., FMEA) is usually underrepresented in the training data and the model may generate more generic answers, lacking the deep domain expertise when reasoning.

\textbf{Complexity and heterogeneity of industrial entities}. Industrial systems often consist of highly complex and heterogeneous assets, each with unique sensors and failure modes. LLMs are expecting to tell the difference between \textit{Compressor fouled} and \textit{Compressor stalled}. Distilling expert knowledge about such varied and intricate concepts into a smaller model is difficult. 

\subsection{Contributions}
To tackle all the challenges above, we propose a knowledge distillation framework for industrial asset health monitoring applications. The paper will cover contributions listed as followed:
 \begin{enumerate}
     \item We present a novel distillation framework designed to semi-automatically transfer Chain-of-Thought reasoning on multi-choice question answering tasks from LLMs to SLMs.
     \item We introduce a novel KnowledgeGraph-inspired method to generate synthetic instructions including pseudo label for industrial domain completely without seed documents.
     \item We perform a thorough qualitative evaluation of in-context learning and fine-tuning using domain knowledge generated by the framework, concluding that the student models achieve substantial performance improvements, ranging from 11\% to 23\%, depending on the base models.
 \end{enumerate}

\section{Related Work}

Chain-of-Thought (CoT) prompting has significantly improved interactions with large language models (LLMs), leading to better results across various datasets, such as MathQA \cite{wei2023chainofthoughtpromptingelicitsreasoning}. This success has inspired research focused on utilizing CoT traces from larger models to distill information, knowledge, and reasoning \cite{mitra2023orca2teachingsmall} \cite{zelikman2022starbootstrappingreasoningreasoning}. The core premise of this research is that information for a target domain can be either pre-existing in the form of question-answer pairs or generated using an additional LLM.

Aligning an LLM to a specific skill has recently emerged as another area of focus. Knowledge generation from a teacher LLM typically begins by leveraging documents or seed instructions \cite{selfinstruct}, \cite{sudalairaj2024lablargescalealignmentchatbots}. More recently, the Magpie-based approach has gained interest, allowing LLMs to generate alignment data in an instruction-free manner \cite{xu2024magpie}. However, extracting domain-specific knowledge from LLMs remains a challenge. Various attempts have been made to generate domain-specific aligned models, such as MediTron-70B for the medical domain \cite{chen2023meditron70bscalingmedicalpretraining} and EntiGraph CPT for long-passage question answering on articles \cite{yang2024syntheticcontinuedpretraining}, among others. These approaches typically rely on large-scale corpora with billions of tokens or begin with a few million tokens to generate additional synthetic data using a teacher model. However, they largely overlook the Industry 4.0 domain. One primary reason for this may be the lack of a qualitatively constructed validation dataset such as other domains PubMedQA \cite{jin-etal-2019-pubmedqa}) and chemical safety (e.g., ChemSafety \cite{zhao2024chemsafetybenchbenchmarkingllmsafety}) as examples.

\section{Methodology}

\begin{figure*}[t]
    \centering
    \includegraphics[width=0.9\linewidth]{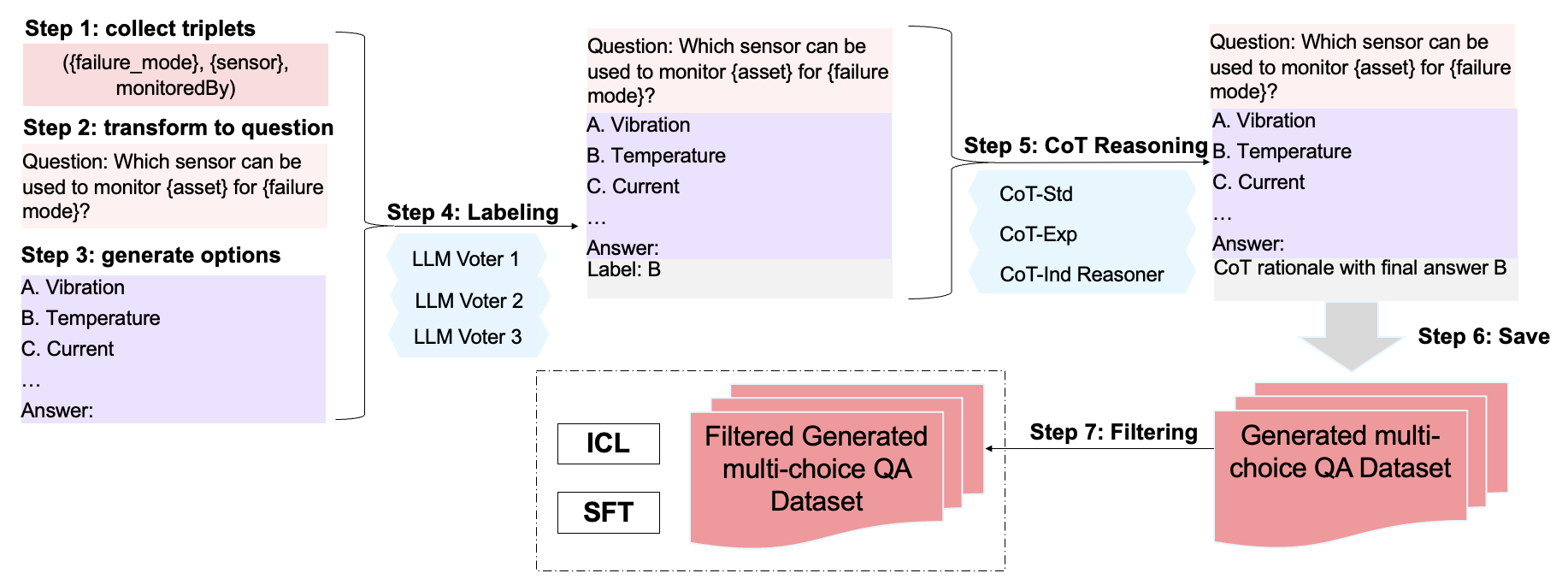}
    \caption{Proposed Knowledge Distillation Framework: Workflow leading to Fine-Tune}
    \label{fig:framework}
\end{figure*}

We present our proposed Knowledge Distillation framework in Figure \ref{fig:framework}, which transfers FMEA knowledge from a teacher model to a student model. The process for generating synthetic multiple-choice question answering (MCQA) data using Chain of Thought (CoT) prompting is described step-by-step in the following sections. Notably, the generation process is seed-free, meaning it does not rely on an initial dataset.

\subsection{KG-based Instruction Generation}
In manufacturing, Knowledge Graphs (KGs) are commonly used to organize relationships and derive domain knowledge \cite{PCA_RDS_2025}. Each edge in a knowledge graph can be represented as a collection of relational triplets \((s, o, r)\), where \(s\) denotes the subject entity, \(o\) represents the object entity, and \(r\) defines the relation between these entities. Within the context of FMEA, three critical relations, as identified by domain experts, are:

\begin{itemize}
    \item \textit{mountedOn}: indicates that a physical \textbf{sensor} is mounted on an industrial \textbf{asset} for the purpose of monitoring or tracking.
    \item \textit{experiencedBy}: indicates that a \textbf{failure mode} is experienced by an industrial \textbf{asset}.
    \item \textit{detectedBy}: indicates that a \textbf{failure mode} can be detected by a \textbf{sensor} associated with the industrial asset.
\end{itemize}

When the subject (\(s\)) or object (\(o\)) of a triplet \(T(s, o, r)\) is omitted, the remaining element becomes a seed for generating LLM instructions. Table \ref{tab:ex_triplet} illustrates an example of how such a seed can be transformed into a question. To facilitate the generation of these questions, we have designed a variety of handcrafted seed templates for each node type. In total, we have 23 distinct seed templates (See Appendix \ref{tab:seed_templates}), which covers all four question categories, as shown in Table \ref{tab:FMEA_questions}.

\begin{table}[h]
    \centering
    \small % Reduce font size for compactness
    \renewcommand{\arraystretch}{1.3} % Adjust row spacing
    \begin{tabular}{p{2cm} p{4.8cm}}    
        \toprule
        \textbf{triplet} (\(T\)) & (\rule{0.6cm}{0.6pt}, Wind Turbine, mountedOn) \\
        \midrule
        \textbf{question} (\(Q\)) & Which sensor is mounted on \underline{Wind Turbine} for monitoring performance of the asset? \\
        \bottomrule
    \end{tabular}
    \caption{Mapping: a seed to a natural language question}
    \label{tab:ex_triplet}
\end{table}

\subsection{Options Generation}
Generating the correct option and multiple distractors is crucial for the effectiveness of a question (\(Q\)). Since the data used is entirely synthetic, we rely on teacher LLMs to generate potential answer choices for a given \(Q\). Let \(U\) represent the universal set of available options in our study, where the content is determined by the type of node, which is omitted at the time of generating \(Q\). We then prompt the LLM to rank candidate options based on their \textbf{correctness criteria} and extract the top \(K\) options from the universal set \(U\) (as specified in the prompt Appendix Table \ref{tab:option_grouping}).  For each of these \(K\) options, we treat each as a correct answer and generate a distinct problem with a slightly rephrased instruction. This approach not only introduces diversity in the correct answers but also reduces the risk of bias by preventing reliance on a single, potentially erroneous, option.

To generate distractors, we retain the bottom \(2K\) options (i.e., less relevant) from the generated response which are those with the least correctness and use them as candidates for the distractors. To further avoid patterns, we randomize the positions of the answers. In summary we generated \(K\) instances of original question \(Q\) with option.  

The use of correctness criteria is our contribution. For example, for a question in Table \ref{tab:ex_triplet}, ``the sensors that can be installed on asset'' is an example of criteria in order to obtain the relevant set of \(K\) options from \(U\). We have defined criteria for each question category in Table \ref{tab:FMEA_questions}. We set \(K\) = 5.

\subsection{Pseudo Ground Truth Labelling}
After constructing the prompt with the instruction, question \(q\), and the options denoted as \textit{A, B, C, D, E, ...}, the LLM is prompted to generate an answer. The labeling process involves selecting the best option from the available choices. We implement a \textbf{majority voting} mechanism using three LLMs (Mixtral Large, Llama-3.1 405b, and ChatGPT) to produce the final answers.

For each multiple-choice question generated, we assign a label if there is clear consensus among the voters. If two voters agree on a particular answer, we evaluate the LLM generated confidence scores and assign a label only if both confidence scores exceed 90. We have used a ``self-guess'' prompt to interact with LLM for generating final answer (Appendix - Table \ref{tab:self-guess}).
 
\subsection{Rationale Generation}
To distill knowledge from Large Language Models (LLMs), we implement Chain of Thought (CoT) based prompting. CoT prompting aids reasoning in multiple-choice question answering by breaking down the process into step-by-step logical steps as well as taking into account the available options. This approach reduces reliance on shallow heuristics and enhances accuracy when tackling complex problems. We selected three variations of trigger statements, as shown in Table \ref{tab:cot_templates}. The question, along with its options, is used with the CoT trigger to generate an LLM-based answer and its rational. If the LLM-generated answer matches the pseudo-label, a rationale is subsequently applied.

\begin{table}[h!]
\centering
\scalebox{0.9}{
\begin{tabular}{|p{1.7cm}|>{\columncolor{blue!10}}p{5.9cm}|}
\hline
\textbf{CoT Stype} & \textbf{Trigger Statement} \\
\hline
\texttt{Standard} & \cellcolor{blue!10} \texttt{Let me think step by step} \\
\hline
\texttt{Inductive} & \cellcolor{blue!10} \texttt{Let me think step by step as a reliability engineer} \\
\hline
\texttt{Expert} & \cellcolor{blue!10} \texttt{Let's use step by step inductive reasoning, given the domain specific nature of the question} \\
\hline
\end{tabular}}
\caption{Chain-Of-Thought Trigger Statements}
\label{tab:cot_templates}
\end{table}

For \textbf{CoT Expert}, the approach mimics expert-level thinking by structuring reasoning around engineering principles and best practices in FMEA. This method aids the model in prioritizing key factors such as failure modes, causes, effects, and detection methods. On the other hand, \textbf{CoT Inductive} allows LLMs to derive conclusions by identifying patterns and relationships within the text. This is particularly useful for handling unfamiliar scenarios or edge cases (e.g., unknown sensors or failure modes), where expert knowledge alone may not suffice. Research suggests that these two CoT variations could potentially outperform \textbf{CoT Standard} in certain situations \cite{liévin2023largelanguagemodelsreason}.

\subsection{Quality Filtering}
We apply several heuristics from \cite{xu2024magpie} to select high-quality generation for down-streaming fine-tuning. Here are the proposed metrics and the empirical thresholds:
\begin{itemize}
    \item Input and Output Length: the total number of characters combining input and output. We filter those generations exceeding max context length of LLMs.
    \item Minimum Neighbor Distance: The embedding distance to the nearest neighbor. Filter the lowest $5\%$ of generations based on scores.
    \item Input Difficulty: LLM-as-a-Judge to determine the difficulty of question on 5 scales (very easy, easy, medium, hard and very hard). Remove very easy generations.
    \item Output Quality: LLM-as-a-Judge to determine the quality of output on 5 scales (very poor, poor, average, good and excellent). Remove very poor and poor generations.

\end{itemize}

Although these quality control methods effectively filter out clearly flawed generations, it remains necessary to assess the quality of the generated data using the teacher model, as discussed in Section \ref{factscore}.

\section{Experiments}
In this section, we conduct extensive experiments to evaluate the effectiveness of the FMEA-related QA data generated by \texttt{FailureSensorIQ} (Appendix: Table \ref{tab:FailureSensorIQ_ex}). \texttt{FailureSensorIQ} is a new dataset we introduced to the community for testing LLM's ability to reason about sensor and failure mode relation. Our goal is to enhance the student model's performance by leveraging knowledge distilled from the teacher model. In other words, we aim to improve SLMs so they can achieve reasoning capabilities comparable to LLMs in the scope of FMEA.

\subsection{Generated Data Statistics}
We use \textbf{Mistral Large} as the teacher model for CoT reasoning generation and rely on \textbf{Mistral Large}, \textbf{Llama-3.1-405B-Instruct}, and \textbf{GPT-4} as the models for majority voting. The choice of teacher models follows prior research findings \cite{constantinides2025failuresensoriqmultichoiceqadataset}. We have collected $54$ candidates for assets, $66$ for sensors, and $59$ for failure modes from ISO standards \cite{iso17359_2018,ISO14224:2016}. For each seed question, we generate $5$ variations (with \(K = 5\)), each featuring different rephrased questions and correct options. By applying three different \textbf{Chain of Thought (CoT)} prompts for reasoning and filtering, the final number of generations is \(6.2k\), \(6.1k\), and \(6.2k\) for \textbf{CoT-Expert}, \textbf{CoT-Inductive}, and \textbf{CoT-Standard}, respectively. We denote the generated datasets as \(D_{\text{gen}}^{\text{CoT-Std}}\), \(D_{\text{gen}}^{\text{CoT-Exp}}\), and \(D_{\text{gen}}^{\text{CoT-Ind}}\). The distribution of assets in the generated data is uneven, with the percentage of each asset ranging from \(1\%\) to \(5\%\).
\subsection{Generated Data Factuality Analysis}  \label{factscore}
To assess the factual consistency of the teacher model-generated data, we conducted an evaluation using an extended version of \textbf{FActScore} \cite{min2023factscorefinegrainedatomicevaluation}, a recent metric that measures the percentage of atomic facts from LLM generations supported by a trusted source, e.g. Wikipedia, Arxiv. We sampled $700$ examples from \(D_{\text{gen}}^{\text{CoT-Exp}}\), ensuring even distribution across $54$ industrial asset types. We run FActScore script with Llama-4-Maverick. 

The results in Table \ref{tab:FActScore} indicate that the teacher-generated data achieves a FActScore of \(70.8\%\), which is slightly higher than the \(69.8\%\) achieved by the FailureSensorIQ benchmark, which means nearly \(70\%\) of facts of our teacher LLM generations are backed by trusted source. Additionally, the responding rate remains high at \(89.6\%\), reinforcing the usability of the generated samples, as the model provides substantive responses rather than abstaining.

\begin{table}[h]
\centering
\setlength{\tabcolsep}{3pt}
\fontsize{10}{12}\selectfont % set font size to 10pt with 12pt line spacing
\begin{tabular}{@{}lcc@{}}
\toprule
 & \textbf{Responding} (\%) & \textbf{FActScore} (\%) \\
\midrule
FailureSensorIQ & 94.7 & 69.8 \\
Generated data & 89.6 & 70.8 \\
\bottomrule
\end{tabular}
\caption{FActScore: Factuality Comparison Between FailureSensorIQ and Teacher Generated Data}
\label{tab:FActScore}
\end{table}

\subsection{Benchmark Dataset}
We select the \textbf{FailureSensorIQ}\footnote{\href{https://huggingface.co/datasets/ibm-research/AssetOpsBench}{https://huggingface.co/datasets/ibm-research/assetopsbench}} dataset \cite{constantinides2025failuresensoriqmultichoiceqadataset} derived and curated from ISO Standards which contains 2667 multi-choice single-true questions with ground truths. The dataset is designed to access the ability to reason and understand the relations between sensor/parameter and failures/faults for assets in Industry 4.0. This dataset covers 10 assets and Table \ref{tab:distribution} lists some distributional information of the dataset. 
\begin{table}[htbp]
\centering
\scalebox{0.9}{
\begin{tabular}{|>{\raggedright\arraybackslash}p{2cm}|p{5.7cm}|}

\hline
\rowcolor{gray!30}  % Light gray background for header row
\textbf{Distribution Type} & \textbf{Distribution Value} \\
\hline
\rowcolor{gray!10}  % Light gray background for first row
Asset Distribution & \textcolor{blue}{Electric Motor} (234), \textcolor{red}{Steam Turbine} (171), \textcolor{orange}{Aero Gas Turbine} (336), \textcolor{purple}{Industrial Gas Turbine} (240), \textcolor{green}{Pump} (152), \textcolor{brown}{Compressor} (220), \textcolor{cyan}{Reciprocating IC Engine} (336), \textcolor{magenta}{Electric Generator} (234), \textcolor{yellow}{Fan} (200), \textcolor{pink}{Power Transformer} (544) \\
\hline
\rowcolor{gray!10}  % Light gray background for second row
Option Distribution & Option A: \textcolor{blue}{752}, Option B: \textcolor{red}{729}, Option C: \textcolor{orange}{491}, Option D: \textcolor{purple}{408}, Option E: \textcolor{green}{208} \\
\hline
\rowcolor{gray!10}  % Light gray background for third row
Distribution of Questions & 2-options: \textcolor{blue}{487}, 3-options: \textcolor{red}{266}, 4-options: \textcolor{orange}{389}, 5-options: \textcolor{purple}{1525} \\
\hline
\end{tabular}}
\caption{Distribution Types and Their Values for FailureSensorIQ MCQA dataset}
\label{tab:distribution}
\end{table}

\subsection{Evaluation Metrics}
Accuracy is the most commonly used metric for tests on MCQA tasks. However recent study \cite{li2024pertevalunveilingrealknowledge} demonstrates a collection of metrics that comprehensively examine the performance of LLMs on MCQA tasks based on response pattern. The proposed metrics are the percentage of LLM responses patterns as followed with denotation: (1) \textbf{$P_{\text{single-correct}}$} - single correct selection (accuracy); (2) \textbf{$P_{\text{single-wrong}}$} - single wrong selection; (3) \textbf{$P_{\text{invalid}}$} - invalid selection (none of answers in the response); (4) \textbf{$P_{\text{mul-correct}}$} - multiple selections with the correct one; (5) \textbf{$P_{\text{mul-wrong}}$} - multiple selections w/o the correct one.

\begin{table*}[t]
    \centering
    \renewcommand{\arraystretch}{1.2} % Adjust row spacing
    \scalebox{0.82}{
    \begin{tabular}{c | c | c | c c c c}
        \toprule
        \multirow{2}{*}{\textbf{LLM}} & \multirow{2}{*}{\textbf{Zero-Shot}} &  \textbf{Few-Shot }& \multicolumn{4}{c}{\textbf{Many-Shot }(generation-based)} \\
        &  & 5 curated & $N=5$ & $N=10$ & $N=20 $ & $N=50$ \\
        \midrule
        Llama-3.1-70B-Instruct  & 249 & 303  & \textbf{316}   & 304  & 313 & 310\\
        Llama-3.1-405B-Instruct  & 251 & 313  & \textbf{317}   & 315  & 316 & \textbf{317}\\
        Mistral-Large-Instruct & 248 & 298  & \textbf{320}   & 315  & 310 & 315\\
        Llama-3.2-90B-Vision-Instruct  & 249 & 300  & 317   & 304  & \textbf{318} & 312 \\ \hline
        Ministral-8B-Instruct & 218 & 245  & 262   & 275  & 275 & \textbf{278}\\
        Llama-3.1-8B-Instruct  & 220 & 277  & 288   & 292  & \textbf{294} & 287\\
        Granite-3.1-8B-Instruct & 187 & 196  & 206   & 206  & 209 & \textbf{210}\\
        \bottomrule
    \end{tabular}}
    \caption{Correctness of LLM inference on 500 FailureSensorIQ questions. This comparison examines the zero-shot baseline against few-shot/many-shot learning, using expert-curated examples vs. examples from $D_{\text{gen}}^{\text{CoT-Exp}}$.}
    \label{tab:icl}
\end{table*}

\subsection{In-Context Learning}  \label{In-Context Learning}
\textbf{In-context learning (ICL)} is a technique that allows Large Language Models (LLMs) to adapt to new tasks during inference by providing a prompt containing task examples. Recently, with the increasing context length of LLMs ($\geq$ 128K), researchers have explored the impact of many-shot learning versus fine-tuning the model \cite{agarwal2024manyshotincontextlearning}. In this experiment, we evaluate the effects of incorporating examples from \(D_{\text{gen}}\) into the prompt when performing inference on the benchmark dataset \textbf{FailureSensorIQ}. We experimented with three options: (1) \textbf{Zero-shot learning}, which tests LLMs with no external knowledge during inference; (2) \textbf{Few-shot learning}, using 5 expert-curated examples. These examples are carefully handpicked to demonstrate the FMEA task, including reasoning processes. The quality of the examples is evaluated by domain experts; (3) \textbf{Many-shot learning} using examples from \(D_{\text{gen}}^{\text{CoT-Exp}}\).

In the case of many-shot learning, we use the \textbf{all-MiniLM-L6-v2} model \cite{allminilm_l6_v2} to compute the embeddings (vector representations) of each question in the synthetic data. During inference, we also convert each query from the benchmark dataset into an embedding using the same embedding model. We utilize cosine similarity to select the top \(N\) relevant generations from \(D_{\text{gen}}^{\text{CoT-Exp}}\) to use as context for the query. Here, \(N \in \{1, 5, 10, 20\}\). It is important to note that the out-of-the-box \textbf{all-MiniLM-L6-v2} is specifically optimized for speed and memory efficiency.

We randomly sample 500 questions from the 2667-question \textbf{FailureSensorIQ} dataset. The number of correct inferences made by LLMs from three open-source LLM families : Llama, Mistral, and Granite, are shown in Table \ref{tab:icl}. The results demonstrate that performance generally improves when transitioning from zero-shot to few-shot learning. \textbf{Llama-3.1-70B-Instruct} shows a substantial increase, reaching 303 correct inferences with 5 curated examples, while \textbf{Mistral-Large-Instruct} achieves 320 correct inferences, maintaining strong performance across various few-shot setups. Larger models tend to consistently outperform smaller models within the same family, primarily due to their larger parameter space. Larger language models also tend to have longer context lengths, which enables them to consume more context and knowledge from additional examples, thereby boosting performance. However, the performance gains begin to plateau when the number of generated examples exceeds 20. Interestingly, larger model tend to perform with less samples whereas the smaller model tend to perform better with more examples.  

The performance improvement from zero-shot to curated few-shot learning is more significant than the improvement from curated few-shot to generation-based few-shot. This is because the model effectively ``learns'' the task when adding curated few-shot examples, whereas the transition from curated few-shot to generation-based few-shot does not introduce a substantial learning step. Additionally, many-shot learning using \(N = 5\) generally outperforms curated samples based example, clearly demonstrating the advantage of in-context learning, where samples are dynamically selected based on the input query.

In conclusion, the generated data from our proposed framework provides a noticeable improvement in contextual understanding during the inference. However, due to model saturation and the noise introduced by potentially low-quality generations, in-context learning may not fully capitalize on the distilled knowledge from the teacher model.

\begin{table*}[ht!]
\centering
\renewcommand{\arraystretch}{1.5} % Adjust row spacing

\scalebox{0.8}{
\begin{tabular}{|c|c|c|c|c|c|c|c|}
\hline
\textbf{Model} & \multicolumn{2}{c|}{\textbf{Experiment Settings}} & \multicolumn{5}{c|}{\textbf{Evaluation Scores}} \\ \hline
(baselines) & \textbf{Synth. Data} & \textbf{Prompting} & \textbf{$P_{\text{single-correct}}$} & \textbf{$P_{\text{invalid}}$} & \textbf{$P_{\text{mul-correct}}$} & \textbf{$P_{\text{single-wrong}}$} & \textbf{$P_{\text{mul-wrong}}$} \\ \hline
\multirow{1}{*}{Llama-3.1-405B-Instruct} & N/A &  CoT Std & \cellcolor{lightgray} 0.5126 & 0.0019 & 0.1691 & 0.258 & 0.0585 \\ \hline
\multirow{1}{*}{Llama-3.1-8B-Instruct} & N/A & direct & 0.4012 & 0.012 & 0.1991 & 0.3048 & 0.0829 \\ \hline
\multirow{1}{*}{Mistral-Large-Instruct} & N/A & direct & 0.5009 & 0.0244 & 0.186 & 0.2295 & 0.0592 \\ \hline
\multirow{1}{*}{Ministral-8B-Instruct} & N/A & direct & 0.264 & 0 & 0.4113 & 0.1894 & 0.1354 \\ \hline
\multirow{1}{*}{Granite-3.1-8B Instruct} & N/A & direct & 0.2411 & 0.0015 & 0.4046 & 0.1916 & 0.1612 \\ \hline \hline
\multirow{1}{*}{FT on} & $D_{\text{gen}}^{\text{CoT-Std}}$ & direct & \cellcolor{lightgray} 0.5111 & 0.0071 & 0.1095 & 0.3116 & 0.0607 \\ \cline{2-8}
\multirow{1}{*}{Llama-3.1-8B-Instruct} & $D_{\text{gen}}^{\text{CoT-Exp}}$ & direct & 0.4698 & 0.0049 & 0.0979 & 0.375 & 0.0525 \\ \cline{2-8}
 & $D_{\text{gen}}^{\text{CoT-Ind}}$ &  CoT Ind & 0.4387 & 0.0165 & 0.1365 & 0.3168 & 0.0915 \\ \hline \hline
\multirow{1}{*}{FT on} & $D_{\text{gen}}^{\text{CoT-Std}}$ & direct & 0.4402 & 0 & 0.144 & 0.3573 & 0.0585 \\  \cline{2-8}
\multirow{1}{*}{Ministral-8B-Instruct} & $D_{\text{gen}}^{\text{CoT-Exp}}$ & direct & 0.4623 & 0.0004 & 0.1301 & 0.3495 & 0.0577 \\ \cline{2-8}
 & $D_{\text{gen}}^{\text{CoT-Ind}}$ & direct & \cellcolor{lightgray} 0.4938 & 0 & 0.1537 & 0.2913 & 0.0611 \\ \hline 
 \hline
\multirow{1}{*}{FT on} & $D_{\text{gen}}^{\text{CoT-Std}}$ &  CoT Std & 0.3813 & 0.0427 & 0.1552 & 0.3142 & 0.1065 \\\cline{2-8}
\multirow{1}{*}{Granite-3.1-8B-Instruct} & $D_{\text{gen}}^{\text{CoT-Exp}}$ & direct & \cellcolor{lightgray} 0.4083 & 0 & 0.2583 & 0.2163 & 0.117 \\ \cline{2-8}
 & $D_{\text{gen}}^{\text{CoT-Ind}}$ &  CoT Std & 0.4062 & 0.006 & 0.1407 & 0.3952 & 0.0519 \\ \hline
\end{tabular}
}

\caption{Evaluation scores of base models and fine-tuned models on FailureSensorIQ dataset. Column Synth. Data represents the generated dataset used for fine-tuning, and Column Prompting shows the best prompting technique with the highest accuracy. "N/A" indicates that the experiment does not involve fine-tuning.}
\label{tab:model_configurations}
\end{table*}

\subsection{Model Fine Tuning}
\label{sec:model-ft}
With Chain of Thought (CoT) knowledge distillation, the goal of fine-tuning the student model is not only to produce accurate predictions but also to internalize the reasoning behind those predictions in asset health monitoring. Our experimental setup uses \textbf{QLoRA} with 4-bit precision for model fine-tuning. The model leverages \textit{FlashAttention2} for efficient attention computation and supports mixed precision training with \textit{bf16} and \textit{tf32}. The maximum sequence length is set to 2048 tokens, and packing is enabled to optimize memory usage. The training runs for 1 epoch, with a batch size of 8 and gradient accumulation over 2 steps. The learning rate is set to \(2.0 \times 10^{-4}\), with a constant learning rate scheduler and a warmup ratio of 0.1 to gradually ramp up the learning rate. Training is conducted on 2 NVIDIA A100 80GB GPUs. We provide additional discussion on the selection of fine-tuning specifications in the Appendix \ref{sec:appendix-ft}.

The three student models focused on in this experiment are all 8B small language models: \textbf{Llama 3.1 8B Instruct}, \textbf{Mistral 8B Instruct}, and \textbf{Granite 3.1 8B Instruct}. We compare the performance of these student models after knowledge distillation with the baseline models, as shown in Table \ref{tab:model_configurations}.

\subsubsection{Baselines}\textbf{Llama-3.1-405B-Instruct} model achieves the highest \(P_{\text{single-correct}}\) (\textbf{0.51}) among the baselines. \textbf{Mistral-Large-Instruct} model also performs well with \(P_{\text{single-correct}} = 0.50\), but it has a higher \(P_{\text{invalid}} = 0.024\), suggesting it generates more invalid responses. Both \textbf{Mistral-8B-Instruct} and \textbf{Granite-3.1-8B-Instruct} perform the worst among the baselines, with much higher \(P_{\text{mul-correct}}\) values, indicating they often predict multiple correct answers rather than providing a single precise answer.

\begin{figure*}[!h]
    \centering
    \includegraphics[width=1\linewidth]{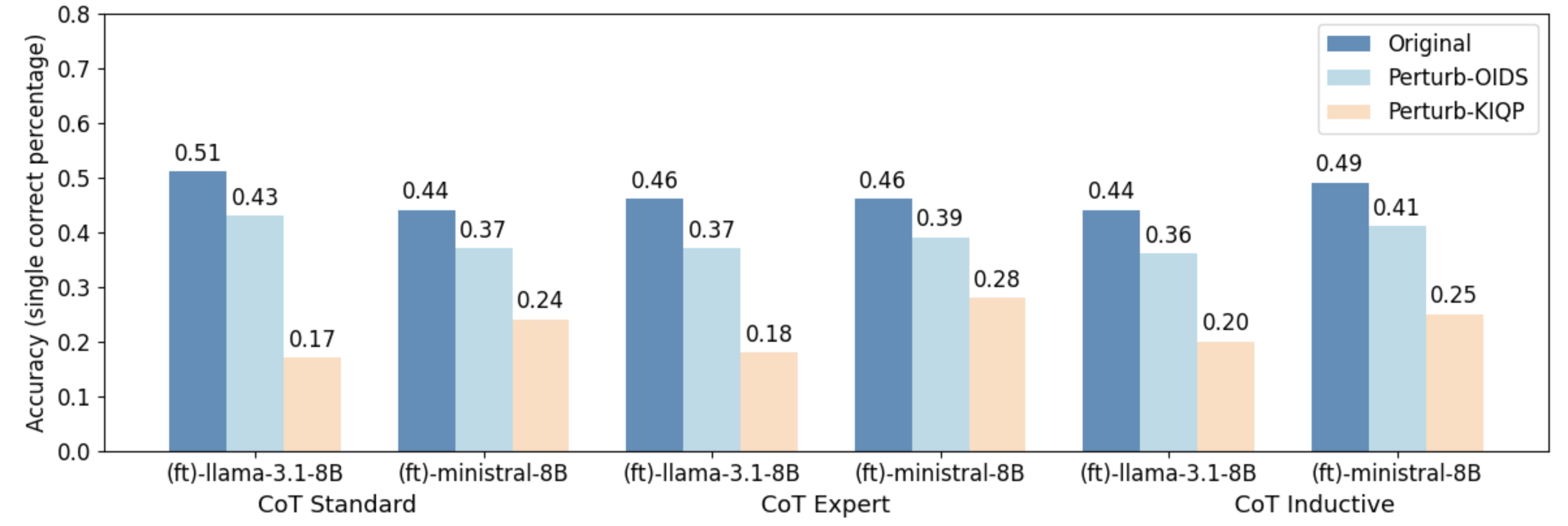}
    \caption{Real knowledge capacity measurement with two knowledge-invariant perturbations: Option ID Shifting (Perturb-OIDS), and Knowledge-Invariant Question Paraphrasing (Perturb-KIQP).}
    \label{fig:perturb_chart}
\end{figure*}

\subsubsection{Impact of Fine-Tuning} In general, fine-tuning on $D_{\text{gen}}^{\text{CoT-Std}}$, $D_{\text{CoT-Exp}}^{\text{CoT-Std}}$, and $D_{\text{gen}}^{\text{CoT-Ind}}$ leads to notable changes in performance across all three student models, with 0.11, 0.23, 0.16  $P_{\text{single-correct}}$ gain respectively. Fine-tuning Llama-3.1-8B-Instruct on $D_{\text{gen}}^{\text{CoT-Std}}$ shows a comparable $P_{\text{single-correct}}$ = 0.51 to Llama-3.1-405B, indicating the SLM, when fine-tuned, achieves a level of performance that is close to that of its larger counterpart. In terms of the choice of CoT style during distilling large models, there is not clear winner among the three. We could argue that CoT-Expert has a slight edge since it has a low change to generate invalid responses, and maintain higher scores in both $P_{\text{single-correct}}$ and $P_{\text{mul-correct}}$. 

\subsubsection{Impact of CoT Fine-Tuning}During inference, we apply prompt engineering to each queries including direct prompting, and three CoT variations listed in Table \ref{tab:cot_templates}.  Then we record the best prompting style with highest $P_{\text{single-correct}}$ in  Column \textit{Prompting} in Table \ref{tab:cot_templates}. We notice that direct prompting is the best one in most experiment settings. This proves the direct prompting is effective enough after student models learn the CoT-style reasoning via knowledge distillation.

\textbf{Observations on Failure Types}. $P_{\text{invalid}}$ remains low across most models, except for the baseline Mistral-Large-Instruct and fine-tuned Llama-3.1-8B-Instruct (CoT-Inductive).  Fine-tuning generally reduces $P_{\text{mul-correct}}$, meaning models are more confident in selecting a single correct answer instead of multiple. $P_{\text{single-wrong}}$  increases for fine-tuned models, suggesting fine-tuning makes models more decisive but also slightly increases the risk of choosing incorrect answers.

\subsection{Ablation Study}
We conducted two ablation studies that isolates the impact of individual components in the framework. Specifically, we evaluate:
    \begin{enumerate}
        \item \textbf{Without rationale} (direct answer only): In this setting, we fine-tuned the student model using only the final answers from the teacher without any accompanying chain-of-thought (CoT) rationales.
        \item \textbf{Incorrect pseudo-labeling} (mismatched answer-rationale pairs): Here, we deliberately introduced noise into the pseudo-labeling process by pairing rationales with incorrect final answers where the rationale and answer do not align.
    \end{enumerate}
We compare the $P_{\text{single-correct}}$ of the two experiments with the baseline (Table \ref{tab:ablation}). All fine-tuning is conducted with QLoRA and the rationale generation is based on CoT Standard. The ablation study results reveal the dramatic performance degradation under incorrect pseudo-labeling, with drops ranging from $13.3\%$ to a severe $21.4\%$. This suggests that students learn not just the reasoning patterns but also the consistency between thinking process and outcome. Interestingly, the impact of removing rationales varies significantly across models: while Llama-3.1-8B shows a moderate $5.2\%$ performance drop without rationales, Ministral-8B actually improves slightly. This heterogeneous response suggests the capacity to effectively utilize CoT reasoning during fine-tuning is model-dependent. For practical applications, these findings suggest practitioners should first evaluate model performance with direct answer fine-tuning before implementing more computationally expensive CoT reasoning approaches.

\begin{table}[h]
\centering

\setlength{\tabcolsep}{2pt}
{\fontsize{10}{12}\selectfont % 12pt font with 14pt line spacing
\begin{tabular}{@{}lccc@{}}
\toprule
\textbf{Model} & \textbf{Baseline} & \textbf{Without} & \textbf{Incorrect} \\
 & & \textbf{Rationale} & \textbf{Pseudo-labeling} \\
\midrule
Llama-3.1-8B & 0.5111 & 0.4593 & 0.3780 \\
Ministral-8B & 0.4402 & 0.5182 & 0.2265 \\
Granite-3.1-8B & 0.3813 & 0.3296 & 0.1429 \\
\bottomrule
\end{tabular}
\caption{$P_{\text{single-correct}}$ of Ablation Studies vs. Baseline}
\label{tab:ablation}
} % end of font size group
\end{table}

\subsection{Perturbation Study}
One significant challenge in evaluating the performance of Large Language Models (LLMs) on multiple-choice question answering (QA) benchmarks is determining how accurately the scores reflect the model's true reasoning ability and knowledge capacity. To address this issue, a recent evaluation framework, PertEval \cite{li2024pertevalunveilingrealknowledge}, introduces a suite of tests that apply various knowledge-invariant perturbations to benchmarks. We utilize two of these perturbations to assess the performance of our fine-tuned student models:

\begin{enumerate} \item \textbf{Option ID Shifting (OIDS)}: This technique substitutes the original option IDs (A/B/C/D/E) with new identifiers (P/Q/R/S/T). OIDS explores potential LLM selection biases in question answering, a phenomenon observed in certain models. By applying this perturbation to the dataset, we can assess whether the choice of option IDs influences the model's performance. \item \textbf{Knowledge-Invariant Question Paraphrasing (KIQP)}: The questions in the original FailureSensorIQ dataset are concise and straightforward. We apply paraphrasing using Llama-3.0-70B to reword these questions while preserving their intended meaning. \end{enumerate}

These two perturbations allow us to evaluate our models on both format and content levels. In Figure \ref{fig:perturb_chart}, we compare the $P_{\text{single-correct}}$ scores of the fine-tuned Llama-3.1-8B-Instruct and Ministral-8B-Instruct models across three datasets: the original FailureSensorIQ, Perturb-OIDS, and Perturb-KIQP. We observe a significant performance drop after applying the perturbations. Specifically, fine-tuned Llama-3.1-8B-Instruct shows greater instability under perturbations, with an average drop of 0.19 per perturbation ($P_{\text{before perturb}} - P_{\text{after perturb}}$), compared to a 0.14 drop for Ministral-8B-Instruct. The KIQP perturbation leads to a more substantial decline in reasoning ability than the OIDS perturbation, indicating that LLMs often rely on memorized patterns and heuristics rather than a deep understanding of the context.

We find that the KIQP perturbation not only alters the questions but also affects the structure of the MCQA task itself, as it causes the options to appear before the question. Additionally, when applying the base model for the same set of experiments, the accuracy of the Perturb-KIQP dataset was notably lower (close to 0.14 for Llama-3.1-8B-Instruct). In summary, even with fine-tuning, small language models (SLMs) improved their grasp on key concepts (e.g., failure modes, sensors). The performance drop across the three types of CoT fine-tuning remains similar, underscoring the challenges LLMs face when the task context is altered.

\section{Conclusion}
We present an innovative knowledge distillation framework designed for asset health monitoring tasks. Our framework generates high-quality synthetic data using LLMs without relying on initial knowledge documents for assets. By fine-tuning small language models (SLMs) with this domain-specific data, the models exhibit reduced hallucination, improved reasoning accuracy, and enhanced consistency in responses. CoT distillation on multi-choice QA tasks further strengthens the SLM's contextual understanding of industrial entities. The low cost of SLM QLoRA fine-tuning (less than 1 hour per experiment, under 4GB adapter for 8B models) makes it a practical solution for adapting models to industry tasks while maintaining scalability and efficiency. Despite these advantages, challenges remain, particularly in handling perturbations, suggesting that future work could focus on perturbation-aware training or incorporating more diverse perturbation scenarios into the synthetic data generation process.

\section*{Limitations}

Our approach leverages a larger teacher model to generate rationales and question-answer pairs for fine-tuning a smaller student model. Given the domain-specific nature of the task, ensuring the factual accuracy and faithfulness of the generated content is essential. However, large-scale human validation is infeasible, and existing automated methods for verifying scientific truthfulness remain limited in both reliability and domain coverage. Consequently, the student model may inherit subtle inaccuracies from the teacher model, particularly in cases involving less-documented or highly specialized knowledge. 

Additionally, this work focuses on three prototypical and high-frequency Failure Mode and Effect Analysis (FMEA) relations : mountedOn, experiencedBy, and detectedBy, as an initial proof of concept to demonstrate the viability of CoT-based distillation in this context. While our framework is modular and readily extensible to accommodate more complex relationships, the current limited relational coverage may affect generalizability to the full FMEA relational space.

Future work should focus on developing more robust, domain-sensitive evaluation techniques for low-resource and high-precision scientific applications, as well as expanding the framework to encompass a broader range of FMEA relationships to enhance the model's comprehensive understanding of complex industrial systems.
\bibliography{custom}

\appendix

\begin{table*}[b]
\centering
\caption{Performance Comparison Across Different Fine-tuning Methods}
\setlength{\tabcolsep}{3pt}
{\fontsize{10}{14}\selectfont % 12pt font with 14pt line spacing
\begin{tabular}{@{}llcccc@{}}
\toprule
\textbf{LLM} & \textbf{Prompting} & \textbf{Full FT:} & \textbf{Full FT:} & \textbf{LoRA:} & \textbf{QLoRA:} \\
 & \textbf{Technique} & \textbf{$P_{\text{single-correct}}$} & \textbf{$P_{\text{invalid}}$} & \textbf{$P_{\text{single-correct}}$} & \textbf{$P_{\text{single-correct}}$} \\
\midrule
\multirow{3}{*}{Llama-3.1-8B} 
 & CoT standard & 0.1005 & 0.4229 & 0.4796 & 0.5111 \\
 & CoT expert & 0.1080 & 0.4421 & 0.4743 & 0.4698 \\
 & CoT inductive & 0.0829 & 0.4631 & 0.4747 & 0.4387 \\
\midrule
\multirow{3}{*}{Ministral-8B} 
 & CoT standard & 0.0937 & 0.5692 & 0.4214 & 0.4402 \\
 & CoT expert & 0.0960 & 0.5816 & 0.5088 & 0.4623 \\
 & CoT inductive & 0.0956 & 0.5681 & 0.4627 & 0.4938 \\
\midrule
\multirow{3}{*}{Granite-3.1-8B} 
 & CoT standard & 0.2343 & 0.0907 & 0.4143 & 0.3813 \\
 & CoT expert & 0.2373 & 0.0900 & 0.4526 & 0.4083 \\
 & CoT inductive & 0.2295 & 0.0885 & 0.4128 & 0.4062 \\
\bottomrule
\end{tabular}
\label{tab:ft-specs}
} % end of font size group
\end{table*}
\section{Appendix}
\label{sec:appendix}
\subsection{Fine Tuning: Full FT vs. LoRA vs. QLoRA}
\label{sec:appendix-ft}
In this section, we compare the performance of three fine-tuning approaches: Full Fine-Tuning (Full FT), Low-Rank Adaptation (LoRA), and Quantized LoRA (QLoRA), on the tasks described in Section \ref{sec:model-ft}. We further discuss the rationale behind ultimately selecting QLoRA.

Our results in Table \ref{tab:ft-specs} show that Full FT frequently degrades performance relative to LoRA, largely due to a high proportion of invalid or incomplete responses. This degradation likely arises from the model overfitting to surface-level patterns in the training data while failing to preserve the intended reasoning behaviors. 

Both LoRA and QLoRA preserve response quality, producing substantially more valid and reasoned outputs, with comparable performance on FailureSensorIQ benchmark tasks. Between the two, LoRA and QLoRA achieve similar accuracy, but QLoRA provides real-world advantages, including a lower memory footprint, faster training time, and more efficient inference.

\begin{table*}[ht!]
\small
\centering
\renewcommand{\arraystretch}{1.5} % Increase row height for better spacing
\begin{tabular}{|>{\columncolor[HTML]{D9EAD3}}m{4cm}|m{10cm}|}
\hline
\multicolumn{2}{|c|}{\cellcolor[HTML]{A9D08E} \textbf{Seed Templates for Sensor and Failure Mode Inquiry}} \\ \hline
\rowcolor[HTML]{A9D08E} 
\textbf{Category} & \textbf{Example Templates} \\ \hline
\multirow{4}{4cm}{\textbf{Sensor}} 
Asset to Sensor & Which sensor could be installed on this asset \{asset\_class\}? \\ \cline{2-2}
& Is there a sensor that can be mounted on this asset \{asset\_class\}? \\ \cline{2-2}
& Can you identify a sensor that could work with this asset \{asset\_class\}? \\ \cline{2-2}
& Which sensor is recommended to track performance and identify anomalies for this asset \{asset\_class\}? \\ \hline
\multirow{10}{4cm}{\textbf{Failure Mode Inquiry Templates}} 
& Which is the most common failure mode associated with the asset \{asset\_class\}? \\ \cline{2-2}
Asset to Failure Mode & Which failure mode should be monitored for the asset \{asset\_class\}? \\ \cline{2-2}
& Which failure mode can occur in the asset \{asset\_class\} during operation? \\ \cline{2-2}
& Which is the failure scenario that the asset \{asset\_class\} might encounter? \\ \cline{2-2}
& Which failure mode is most likely to occur with the asset \{asset\_class\}? \\ \hline
Sensor to Failure Mode& In the context of \{asset\_class\}, which failure mode is most relevant when \{relevant\_sensor\} shows abnormal readings? \\ \cline{2-2}
& Which is the most relevant failure mode for \{asset\_class\} if \{relevant\_sensor\} exhibits abnormal readings? \\ \cline{2-2}
& Which failure mode should be considered for \{asset\_class\} when abnormal readings are detected by \{relevant\_sensor\}? \\ \cline{2-2}
& When \{relevant\_sensor\} in \{asset\_class\} displays abnormal readings, which failure mode is the most applicable? \\ \cline{2-2}
& For \{asset\_class\}, what is the key failure mode when \{relevant\_sensor\} has abnormal readings? \\ \cline{2-2}
& What is the most likely failure mode for \{asset\_class\} when \{relevant\_sensor\} indicates abnormal behavior? \\ \hline

\multirow{8}{4cm}{\textbf{Sensor Selection for Failure Mode Templates}} 
Failure Mode to Sensor & Which sensor can be used to monitor asset \{asset\_class\} for failure mode \{relevant\_failure\_mode\}? \\ \cline{2-2}
& What sensor is suitable for monitoring \{asset\_class\} to detect \{relevant\_failure\_mode\}? \\ \cline{2-2}
& What sensor can be utilized to monitor \{asset\_class\} for signs of \{relevant\_failure\_mode\}? \\ \cline{2-2}
& Which sensor is best suited to monitor \{asset\_class\} for the occurrence of \{relevant\_failure\_mode\}? \\ \cline{2-2}
& In an \{asset\_class\}, which sensor is designed to track \{relevant\_failure\_mode\}? \\ \cline{2-2}
& In the context of \{asset\_class\}, which sensor can help in identifying \{relevant\_failure\_mode\}? \\ \cline{2-2}
& Which sensor would you recommend for monitoring \{asset\_class\} to detect \{relevant\_failure\_mode\}? \\ \cline{2-2}
& Which sensor can effectively monitor \{asset\_class\} for potential \{relevant\_failure\_mode\}? \\ \hline
\end{tabular}
\caption{Template for Question Generation}
\label{tab:seed_templates}
\end{table*}

\begin{table*}[h!]
\centering
\small
\begin{tabular}{|p{2cm}|>{\columncolor{blue!10}}p{10cm}|}
\hline
\texttt{Question template} & \cellcolor{blue!10} \texttt{Divide the following choices into two groups. First group is \textcolor{red}{\{relevance criteria\}}. Second group is \textcolor{red}{\{irrelevance criteria\}}.
Here are a list of choices: \textcolor{red}{\{choices\}}.
Output the first group in the first line. Output the second group in the second line. Format of the output should be:
\newline First group: ["choice1", "choice2", "choice3", ...]
\newline Second group: ["choice4", "choice5", "choice6", ...]} \\
\hline
\texttt{Relevance criteria} & \cellcolor{blue!10} \texttt{failure modes that are the most common failure modes associated with \textcolor{red}{\{asset class\}} sorted by relevancy}\\
\hline
\texttt{Irrelevance criteria} & \cellcolor{blue!10} \texttt{the failure modes that are most unlikely to occur with \textcolor{red}{\{asset class\}} sorted by unlikelihood} \\
\hline
\texttt{Choices} & \cellcolor{blue!10} \texttt{fail to start, failure to stop, ..., bearing wear, unbalance, ...} \\
\hline
\end{tabular}
\caption{Example of question template that groups and ranks the options in options generation process.}
\label{tab:option_grouping}
\end{table*}

\begin{table*}[h!]
\centering
\small
\begin{tabular}{|>{\columncolor{blue!10}}p{10cm}|}
\hline
 \cellcolor{blue!10} \texttt{Here is the question:
\newline Question: \textcolor{red}{\{question\}}
\newline \textcolor{red}{\{options\}}
\newline \newline Please provide your best guess for the answer to the following question and include a confidence score between 0 to 100, an explanation, and a rationale for your answer in the following JSON format:
\newline ```json
\newline \{
\newline "answer": "Your answer here", 
\newline "explanation": "Your explanation here", 
\newline "confidence\_score": "Your score here", 
\newline "rationale": "Your answer here", 
\newline \}
\newline ```
} \\
\hline
\end{tabular}
\caption{Self-guess prompting to extract confidence score and rationale from the response}
\label{tab:self-guess}
\end{table*}

\begin{table*}[h!]
\centering
\small
\begin{tabular}{|p{2cm}|>{\columncolor{blue!10}}p{10cm}|}
\hline
\texttt{Failure Mode to Sensor} & \cellcolor{blue!10} \texttt{For electric motor, if a failure event rotor windings fault occurs, which sensor out of the choices is the most relevant sensor regarding the occurrence of the failure event?
\newline 
\newline A. partial discharge
\newline B. resistance
\newline C. oil debris
\newline D. current
\newline E. voltage
\newline 
\newline Answer: \textcolor{red}{D}} \\
\hline
\texttt{Sensor to Failure Mode} & \cellcolor{blue!10} \texttt{Which failure mode is most relevant for steam turbine if there are abnormal readings from coast down time?
\newline 
\newline A. unequal expansion
\newline B. misalignment
\newline C. bearing damage
\newline D. unbalance
\newline E. damaged labyrinth
\newline 
\newline Answer: \textcolor{red}{C} } \\
\hline
\end{tabular}
\caption{Examples of FailureSensorIQ: a multi-choice question-answering dataset for failure mode and sensor relations}
\label{tab:FailureSensorIQ_ex}
\end{table*}

\end{document}